\documentclass{article} 

\PassOptionsToPackage{numbers}{natbib}
\usepackage{iclr2025_conference,times}

\usepackage{amsmath,amsfonts,bm}

\def\eqref#1{equation~\ref{#1}}

\def\1{\bm{1}}

\DeclareMathAlphabet{\mathsfit}{\encodingdefault}{\sfdefault}{m}{sl}
\SetMathAlphabet{\mathsfit}{bold}{\encodingdefault}{\sfdefault}{bx}{n}

\usepackage{hyperref}
\usepackage{url}

\usepackage{multirow}

\usepackage{amssymb}
\usepackage{enumitem}
\usepackage{placeins}

\usepackage{amsmath}
\usepackage{booktabs}
\usepackage{svg}
\usepackage{mathtools}

\usepackage{pifont} 
\usepackage{tcolorbox} 
\usepackage{graphicx}
\usepackage{geometry}

\usepackage{longtable}
\usepackage{caption}
\usepackage{array}  

\usepackage{multicol}

\title{Unreflected Use of Tabular Data Repositories Can Undermine Research Quality}

\usepackage{authblk}

\makeatletter
\renewcommand\AB@affilsepx{\quad} 
\makeatother

\author[1]{Andrej Tschalzev}
\author[2]{Lennart Purucker}
\author[3]{Stefan Lüdtke}
\author[4,5,2]{\\Frank Hutter}
\author[6]{Christian Bartelt}
\author[1]{Heiner Stuckenschmidt}
\affil[1]{University of Mannheim}
\affil[2]{University of Freiburg}
\makeatletter
\renewcommand\AB@affilsepx{\\} 
\makeatother
\affil[3]{University of Rostock}
\makeatletter
\renewcommand\AB@affilsepx{\quad} 
\makeatother
\affil[4]{ELLIS Institute Tübingen}
\affil[5]{PriorLabs}
\affil[6]{Clausthal University of Technology}

\iclrfinalcopy 
\begin{document}

\maketitle

\begin{abstract}
Data repositories have accumulated a large number of tabular datasets from various domains. Machine Learning researchers are actively using these datasets to evaluate novel approaches. Consequently, data repositories have an important standing in tabular data research. They not only host datasets but also provide information on how to use them in supervised learning tasks. In this paper, we argue that, despite great achievements in usability, the unreflected usage of datasets from data repositories may have led to reduced research quality and scientific rigor. 
We present examples from prominent recent studies that illustrate the problematic use of datasets from OpenML, a large data repository for tabular data.
Our illustrations help users of data repositories avoid falling into the traps of (1) using suboptimal model selection strategies, (2) overlooking strong baselines, and (3) inappropriate preprocessing.
In response, we discuss possible solutions for how data repositories can prevent the inappropriate use of datasets and become the cornerstones for improved overall quality of empirical research studies.
\end{abstract}

\section{Introduction}


Public tabular data repositories such as the UCI ML Repository\footnote{\url{https://archive.ics.uci.edu/}}, OpenML\footnote{\url{https://openml.org/}~\citep{vanschoren2014openml}}, Kaggle\footnote{\url{https://www.kaggle.com/datasets}}, and Hugging Face\footnote{\url{https://huggingface.co/datasets}} are key institutions shaping which and how datasets are used in research \citep{longjohnbenchmark}.
In tabular data research, the OpenML repository is used extensively \citep{gijsbers2019open,tabrepo2024,liu2024talent,hollmann2025accurate}. 
Community members, including OpenML maintainers, curated suites of datasets for benchmarking, such as OpenML-CC18~\citep{bischl2017openml}, the AutoML Benchmark~\citep{gijsbers2024amlb}, 
the \cite{grinsztajn2022tree} benchmark, and TabZilla~\citep{mcelfresh2023neural}. 

A driving factor for tabular data repository usage is the recent increase in efforts to transfer the success of deep learning to the tabular domain. 
The development of novel neural network models \citep{arik2021tabnet,chang2021node,gorishniy2021revisiting,gorishniy2023tabr,gorishniy2024tabm}, and more recently tabular foundation models \citep{gardner2024benchmarking,hollmann2025accurate} dominates the tabular machine learning community. 
In response, recent comparative studies try to gather as many datasets as possible to facilitate a rigorous and comprehensive evaluation of novel approaches \citep{grinsztajn2022tree,mcelfresh2023neural,ye2024closer}. While \cite{mcelfresh2023neural} used 196 datasets, a recent study scales up to 300 datasets from OpenML \citep{ye2024closer}. Similarly, studies evaluating foundation models seem to include as many datasets from these benchmarks as possible, apparently taking their quality and appropriateness for granted \citep{yan2024making, gardner2024benchmarking}. 

Different authors have recently criticized the intense focus on model development and the limited attention to data quality.
Existing benchmarks often use outdated data \citep{kohli2024towards}, ignore task-specific preprocessing \citep{tschalzev2024data}, or use inappropriate data splits \citep{rubachev2024tabred}. 
Evidently, there is increased skepticism regarding the use of data in tabular data research. In this paper, we demonstrate that, although most data repositories are open-upload communities with limited data quality assurance, users tend to use the datasets without reflection—that is, without critically inspecting the available data or task.
We hypothesize that this stems from a limited awareness of consequences for data misusage among researchers as well as design choices made by data repositories. 
The role of data repositories in machine learning research has recently gained attention as a topic of study \citep{longjohnbenchmark}. However, their specific impact on tabular data research has received limited attention \citep{rubachev2024tabred}, despite tabular data repositories being among the oldest and most used \citep{vanschoren2014openml,benjelloun2020google}. 

\paragraph{Our contributions.}
We focus our analysis on studies that seek to assess or advance the state of the art in supervised machine learning for tabular data, a frequently stated objective in recent literature \citep{grinsztajn2022tree,mcelfresh2023neural,ye2024modern,gorishniy2024tabm}. We promote a more reflective use of tabular data repositories by (\textbf{1}) highlighting three critical issues in tabular benchmarking and (\textbf{2}) helping researchers understand and avoid related mistakes. Furthermore, we (\textbf{3}) outline how data repositories can help prevent these issues and serve as institutions that provide guidelines for the appropriate use of tabular data.
In particular, we show that:
\begin{enumerate}[labelindent=5em, leftmargin=*] 
    \item [Section \ref{sec:eval_strat}:] Different datasets require distinct validation strategies for model selection, yet these strategies are often absent from the evaluation designs of most studies and data repositories.
    \item [Section \ref{sec:sota}:] Although the same datasets have been used for years, data repositories do not provide baseline performance for each dataset, leading to a fluctuating state-of-the-art across studies.
    \item [Section \ref{sec:preprocessing}:] Results are heavily influenced by which features of a dataset are used and their preprocessing, yet data repositories offer only limited guidance on preprocessing to users.
\end{enumerate}

We use concrete examples from the OpenML repository and recent studies to point out the impact of specific issues. 
Our findings highlight that research users of data repositories tend to use the data and evaluation setups provided by repositories \emph{without critical reflection}. 
This stresses the large impact data repositories have on the standards for evaluation quality.

\section{Effects of Unreflected Data Repository Use on Recent Benchmarks} \label{sec:main_exp}

We focus on two benchmark studies: (1) The TabZilla-hard benchmark \citep{mcelfresh2023neural}, comprising 36 minimally preprocessed datasets selected to expose performance disparities among models. (2) The \citet{grinsztajn2022tree} benchmark, that was curated using data-driven criteria and simplified by removing side issues such as high-cardinality categorical features and missing values. We select these studies for several reasons: 
First, both are recent yet highly influential in tabular data research as the proposed benchmarks are actively used in recent studies. 
Second, the benchmarks make fundamentally different design choices and, therefore, are suitable to highlight different aspects of our scope.
Finally, we acknowledge the significant effort and exceptional attention to experimental details demonstrated in both studies. However, as we will show, even these exceptional studies have flaws in dataset treatment and evaluation design.

\textbf{Experimental Design}  \hspace{1.5mm} For both benchmarks, we train the following models on all datasets: LightGBM \citep{ke2017lightgbm}, XGBoost \cite{chen2016xgboost}, simple MLPs as introduced by \cite{gorishniy2021revisiting}, and TabM \citep{gorishniy2024tabm}. The first three represent commonly used baselines, while TabM represents one of the most recent contributions to tabular deep learning. 
For the TabZilla datasets, we use the train/validation/test splits provided by the benchmark. 
For the Grinsztajn benchmark, we follow the described procedure to obtain train/val/test splits and repeat it 15 times with different random seeds for each dataset to better account for random variance in the experiments. 
For each fold/split, the hyperparameters of each model are independently tuned over 100 trials. 
In line with related work, the best model is selected on the validation data. 
For the TabZilla datasets, we additionally evaluate an alternative model selection procedure where, instead of the default holdout validation split, 5-fold cross-validation (5CV) is used.
In this regime, we obtain predictions by averaging the predictions of all fold models, akin to cross-validation ensembles \citep{krogh1994neural,erickson2020autogluon}. 
See Appendix~\ref{sec:experimental_details} for more experimental details.

\subsection{Inappropriate Validation and Model Selection Strategy Per-Dataset} \label{sec:eval_strat}


\begin{figure}[t]
    \centering
    \includegraphics[width=0.99\columnwidth]{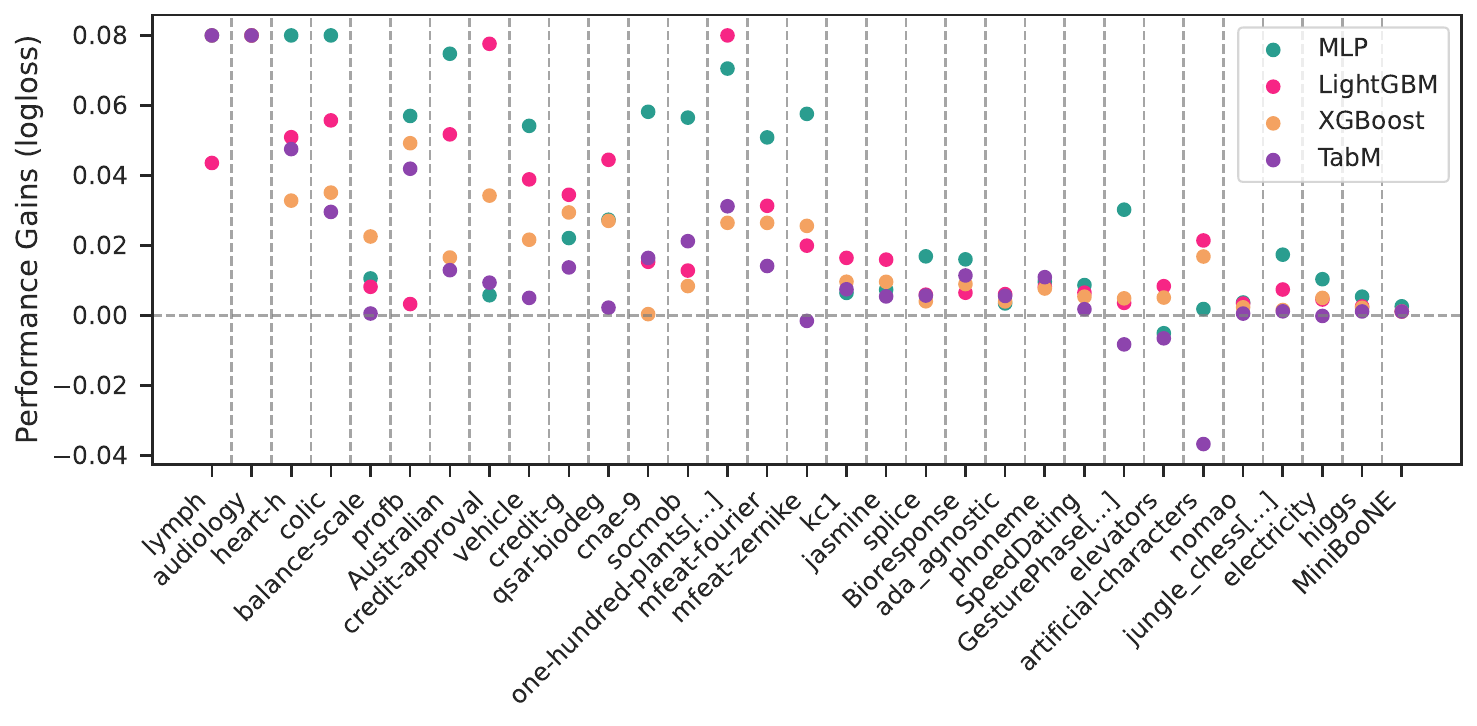}
    \caption{\textbf{5-fold cross-validation (5CV) consistently improves performance over holdout validation}. 
    We show the performance gains obtained by switching from holdout validation to 5CV for the TabZilla benchmark. Each model is compared to itself after 100 trials in both model selection protocols. 
    For each dataset, the zero line corresponds to the model's performance with holdout selection.
     Positive values mean better performance with the 5CV protocol.
    The performance gains are capped at 0.08.
    The datasets are sorted by sample size in ascending order. 
    }
    \label{fig:5cv_gains}
\end{figure}


It is well known that hyperparameter optimization (HPO) is crucial for an accurate evaluation of tabular data tasks \citep{kadra2021well,gorishniy2021revisiting,shwartz2022tabular,tschalzev2024data}. 
However, in HPO, a validation criterion must be defined to select the best model.
While most benchmarks provide train/test splits, validation strategies are surprisingly rarely discussed. Studies often define their own validation strategies, with single-holdout validation as the most commonly used technique for model selection. 
Nevertheless, when the goal is robustly evaluating a model, holdout selection is often not the best choice as it can lead to overfitting the validation criterion \citep{cawley2010over,nagler2024reshuffling}. 
Furthermore, the widespread use of holdout validation in academic studies contrasts with real-world practice, where applying a model to test data after validating only on one single, often small, set of validation samples would be too risky. 
Instead, as can be seen in almost every Kaggle competition\footnote{ \url{https://www.kaggle.com/code/rhysie/learnings-from-the-typical-tabular-pipeline}}, cross-validation is used for model selection - with a varying strategy depending on the task. 

To demonstrate why using single holdout validation is a suboptimal choice, Figure \ref{fig:5cv_gains} illustrates the performance gains of 5-fold cross-validation (5CV) model selection over the holdout method on the TabZilla benchmark. It can be seen that the 5CV method is almost always superior to the holdout method, often yielding strong performance gains for all models. 
While the effect appears stronger for small datasets, many larger datasets are also affected, particularly when using MLPs.
Note that we excluded one dataset because most models already performed flawlessly with holdout selection and four large datasets because 100 trials with 5-fold cross-validation were infeasible for them. See Appendix~\ref{ssec:reproduce} for further explanation.

\begin{figure}[t]
    \centering
    \includegraphics[width=0.99\columnwidth]{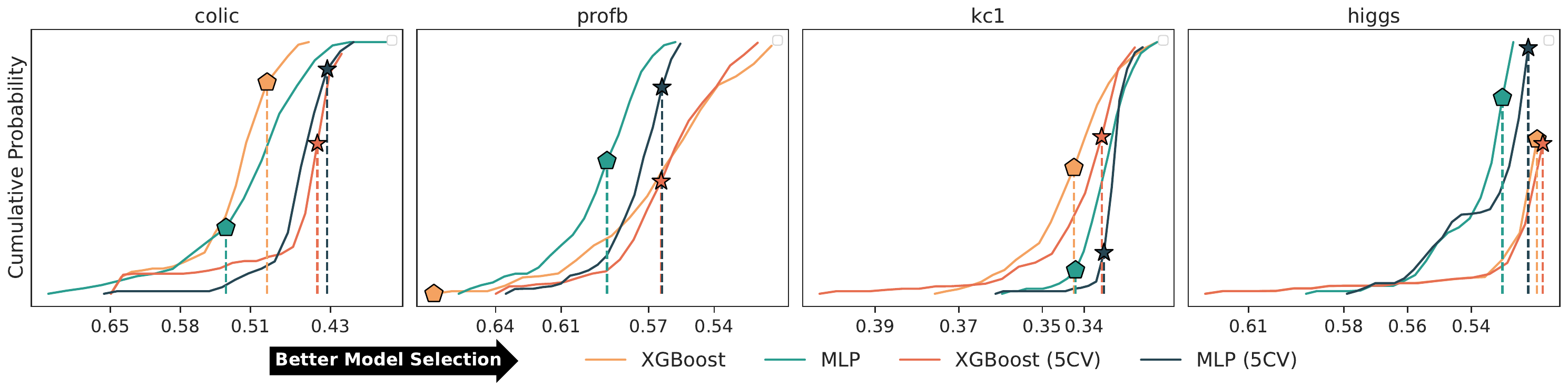}
    \caption{\textbf{Holdout validation is prone to make model selection miss the highest achievable performance.} We show the cumulative density functions of test logloss performance over 100 trials for MLPs and XGBoost. One fold is displayed for each dataset. 
    Stars with vertical lines denote the model selected based on the best validation performance. A position closer to the right on the x-axis means better absolute logloss performance. Two models with the same value on the x-axis mean equal performance. Steeper ascents on the y axis represent dense regions with more trials achieving similar performance. }
    \label{fig:four_examples_valoverfitting}
\end{figure}

Next, we present concrete examples illustrating the impact of an inappropriate validation strategy on the conclusions drawn from model comparisons.
Figure \ref{fig:four_examples_valoverfitting} shows that with the holdout validation strategy, it is more likely to miss the hyperparameters with stronger performance. 
On \textbf{colic}, both models could perform similarly, but the performance of MLPs is estimated to be worse than XGBoost with holdout validation; 
on \textbf{profb}, this behavior is reversed with XGBoost being wrongly estimated to be worse than MLPs; on \textbf{kc1}, the comparison was already correct in the holdout regime, but the probability mass for stronger hyperparameters is higher with 5CV, showing that a negatively biased comparison becomes less likely; lastly, on \textbf{higgs}, the performance gap narrows with 5CV model selection, demonstrating that the issue also affects larger datasets with up to 100,000 samples. 

We emphasize that 5CV does not always solve the issue of overfitting to validation data entirely. Especially for small datasets, other strategies can be more suitable \citep{dietterich1998approximate,raschka2018model,schulz2024constructing}. Nevertheless, given the consistent, often large, performance gains with the simple 5CV method, it can be concluded that the holdout selection method is inappropriate for most datasets in the TabZilla benchmark. Furthermore, since the 36 benchmark datasets were selected from a larger set of 196 datasets based on results obtained with the holdout method, the validity of the entire benchmark may be compromised. 


\textbf{How can researchers avoid this issue?}  \hspace{1.5mm}
Our results show that for most datasets, holdout selection is biased. Hence, researchers should avoid blindly using holdout validation as the default validation strategy.
Furthermore, researchers should investigate whether their models are overfitted to the validation data, i.e., by comparing validation and test performance. If HPO is performed, an important ablation study can be to compare the selected trial to the best trial on the test data, or to visualize the performance distribution over trials as in our study. Note that peeking at the alternative test scenarios is only possible in an academic evaluation context and should only be done for debugging to avoid bias in developing methods.
It is important to note that further research is needed to determine the appropriate validation strategy for different datasets, as no agreed-upon standard exists and automated procedures could be beneficial.

\textbf{How can data repositories prevent this issue?}   \hspace{1.5mm}
Currently, benchmark studies like TabZilla specify validation splits in their code.
In general, validation should remain flexible for users, as different research questions may require distinct validation strategies. 
However, researchers aiming to improve the state of the art often rely on the holdout validation method, even though it frequently leads to underestimating models, thereby preventing robust comparisons.
We argue that data repositories can play a crucial role in addressing this widespread misconception. We suggest that, in addition to specifying the estimation procedure, data repositories should establish a default validation procedure as part of a standardized task schema\footnote{For example, adding a validation procedure to OpenML tasks: \url{https://openml.github.io/openml-python/main/generated/openml.tasks.OpenMLTask.html}}, thereby enabling a flexible and less biased evaluation across datasets.
Sections \ref{sec:sota} and \ref{sec:potential} expand on this suggestion.

\subsection{Obscured State-of-the-art Due to Missing Objective Baseline} \label{sec:sota}
In this subsection, we highlight how the availability of a publicly shared strong baseline for each dataset could have prevented false design choices and misleading conclusions in the TabZilla benchmark study \citep{mcelfresh2023neural} and subsequent studies that build upon it.
We emphasize that we do not use this benchmark study as an example because we consider it inferior to other studies. On the contrary, we greatly appreciate the authors' efforts to provide an extensive evaluation repository to the community. We believe the highlighted issues are systemic and likely affect many other studies.


\begin{table}[tb]
    \caption{\textbf{Better-tuned MLPs outperform all results from the TabZilla benchmark.} We show the performance of algorithms across the TabZilla-hard benchmark suite with improved HPO protocols for XGBoost and MLP. The remaining results are taken from \cite{mcelfresh2023neural}.}
    \centering
    \small
    \begin{tabular}{lrrr}
    \toprule
     & Avg. Rank & Avg. norm. logloss & Avg. logloss \\
    \midrule
    XGBoost (ours, 5CV) & 1.77 & 0.03 & 0.34 \\
    MLP (ours, 5CV) & 2.1 & 0.08 & 0.34 \\
    XGBoost (ours, holdout) & 4.13 & 0.06 & 0.36 \\
    XGBoost & 5.56 & 0.1 & 0.39 \\
    CatBoost & 5.84 & 0.12 & 0.45 \\
    MLP (ours, holdout) & 6.09 & 0.15 & 0.4 \\
    LightGBM & 6.85 & 0.17 & 0.45 \\
    ResNet & 8.12 & 0.22 & 0.49 \\
    SAINT & 8.77 & 0.23 & 0.52 \\
    ... & & & \\
    MLP & 10.79 & 0.39 & 0.96 \\
    ... & & & \\
    KNN & 15.68 & 0.71 & 0.88 \\
    \bottomrule
    \end{tabular}
    \label{tab:validity}
\end{table}

In their experiments for constructing the TabZilla benchmark, \citet{mcelfresh2023neural} used very restrictive HPO time budgets and VRAM management. 
As a result, not all models were trained to their full potential.
To demonstrate the consequences, we train XGBoost and MLPs using the same datasets and train/test splits with our improved HPO protocols. We do not change the models themselves, add our results to the original ones, and repeat the same evaluation as the original study.
Table~\ref{tab:validity} shows that a simple MLP outperforms all 18 models from the TabZilla benchmark only by implementing a more appropriate model selection strategy with 100 HPO trials and 5CV for validation.
While XGBoost is still better on average than MLPs, the performance of both methods was severely underestimated, with MLPs being affected much more strongly. 
Note that we excluded TabPFN, as it was only applicable to 12 datasets in the original study.

\citet{mcelfresh2023neural} analyzed the reasons behind performance differences between neural networks and tree-based models. To do so, they used dataset statistics as meta-features and computed their correlations with the difference in log-loss performance between models.
However, since the performance of the models was systematically underestimated due to the restricted evaluation design, the conclusions of the TabZilla study regarding performance differences are unreliable.
To demonstrate this, Table \ref{tab:corr} shows that the study's findings regarding the most important meta-features do not hold under improved HPO protocols.


As \citet{mcelfresh2023neural} published their full hyperparameter search results, other researchers compared their results \textbf{only} to the precomputed TabZilla benchmark \citep{mueller2024mothernet,thomas2024retrieval,breejen2024context,ma2024tabdpt,xu2024mixture}. 
Our findings demonstrate that improvements over TabZilla could have been achieved even without novel contributions to modeling.
Consequently, studies that relied on the precomputed TabZilla benchmark were misled by an underestimated evaluation baseline, likely impacting their conclusions.

Finally, we argue that a strong, objective per-dataset baseline, independent of individual benchmarking suites or studies, could have prevented many studies from evaluating against suboptimal baselines.
The TabZilla benchmark reuses the same OpenML tasks with train/test splits that have been used for years in the CC-18 benchmark \citep{bischl2017openml}. 
We did not employ novel techniques to obtain the results in Table \ref{tab:validity}. Therefore, it is highly likely that similar performance on the same tasks has already been achieved in other studies.
However, no established standard exists in the research community for systematically tracking the actual state-of-the-art performance of these datasets. 

\begin{table}[tb]
    \caption{\textbf{Conclusions regarding model differences can be compromised by the evaluation design.} We show the meta-features with the highest correlation to the performance difference between the best neural network and the best tree-based model. 
    Regarding the results for \cite{mcelfresh2023neural}, those reported for 196 datasets correspond to Table 14 in their paper. The results for 36 datasets are based on a replication of the same analysis using the published HPO search results for the 36 datasets selected for the benchmark. 
    Finally, we present our results using better-tuned MLPs and XGBoost instead of the original versions. 
    While most of the previously identified trends become even more pronounced for the subset of 36 datasets under the original evaluation design, no strong patterns persist after applying our advanced evaluation design.}
    \centering
    \small
    \begin{tabular}{lcccc}
    \toprule
     Description & \multicolumn{2}{c}{\cite{mcelfresh2023neural}} &\multicolumn{2}{c}{ Ours} \\
     \multicolumn{1}{r}{No. of datasets}& 196 datasets & 36 datasets & 36 datasets & 36 datasets\\
     \multicolumn{1}{r}{No. of trials, validation}  & \multicolumn{2}{c}{$\leq$ 30, holdout} & 100, holdout & 100, 5CV\\
    
    \midrule
    Log number of instances. & 0.63 & 0.42 & 0.05 & 0.22 \\
    Ratio of the size of the data set to the number of features. & 0.55 & 0.73 & -0.06 & -0.02 \\
    Log of the median canonical correlation between each & -0.41 & -0.9 & -0.14 & -0.16 \\
    \ \ \ \ feature and the target. & & & \\
    Log of the min. target class frequency. & -0.35 & -0.85 & -0.0 & 0.18 \\

    \bottomrule
    \end{tabular}
    \label{tab:corr}
\end{table}

\textbf{How can researchers avoid this issue?} \hspace{1.5mm}
We suggest that researchers always include strong baselines and avoid comparing \textit{solely} to precomputed results from prior work.
For large-scale evaluations where not every model can be exhaustively tuned, we suggest using a single strong baseline as a reference. 
Specifically, we suggest 100 trials of hyperparameter optimization with LightGBM without a time limit and with 5-fold cross-validation for model selection. 
Thereby, we suggest using a predefined search space used in practice, such as the search space from AutoML systems like AutoGluon~\citep{erickson2020autogluon}, LightAutoML~\citep{vakhrushev2021lightautoml}, or FLAML~\citep{wang2021flaml}. 
Although this baseline may not achieve the highest possible performance on every dataset, it is unlikely to be far off and is feasible for most studies due to LightGBM's excellent performance and training time tradeoff.

\textbf{How can data repositories prevent this issue?} \hspace{1.5mm}
Although OpenML provides a leaderboard of submitted runs for each task, for most of the available datasets, the last entries date back several years ago\footnote{For example, \url{https://www.openml.org/t/31}, or \url{https://www.openml.org/t/14}}. 
Evidently, the research community does not rely on the current OpenML leaderboard system, so it cannot serve as a performance reference.
Therefore, we suggest that OpenML should enable users to share the performance of a strong baseline as part of the dataset schema or task schema. 
This would also prompt the user to consider an appropriate metric and model selection procedure. 
Displaying the strong baseline performance on OpenML can provide a reference for researchers and reviewers to more quickly rule out contributions that mistakenly claim to achieve improvements over the state-of-the-art. 
For a reference performance on a dataset, an appropriate default task would be required. This aspect will be further discussed in Section \ref{sec:potential}.

\subsection{Insufficient Information for Task-Relevant Preprocessing}  \label{sec:preprocessing}

In this subsection, we build on the results of \citet{rubachev2024tabred} to show how solely relying on dataset statistics for dataset selection has compromised the benchmark of \cite{grinsztajn2022tree}. In particular, we show how using forbidden features led to target leaks, and how the wrong preprocessing led to underestimated model performance.

\begin{figure}[t]
    \centering
    \includegraphics[width=0.99\columnwidth]{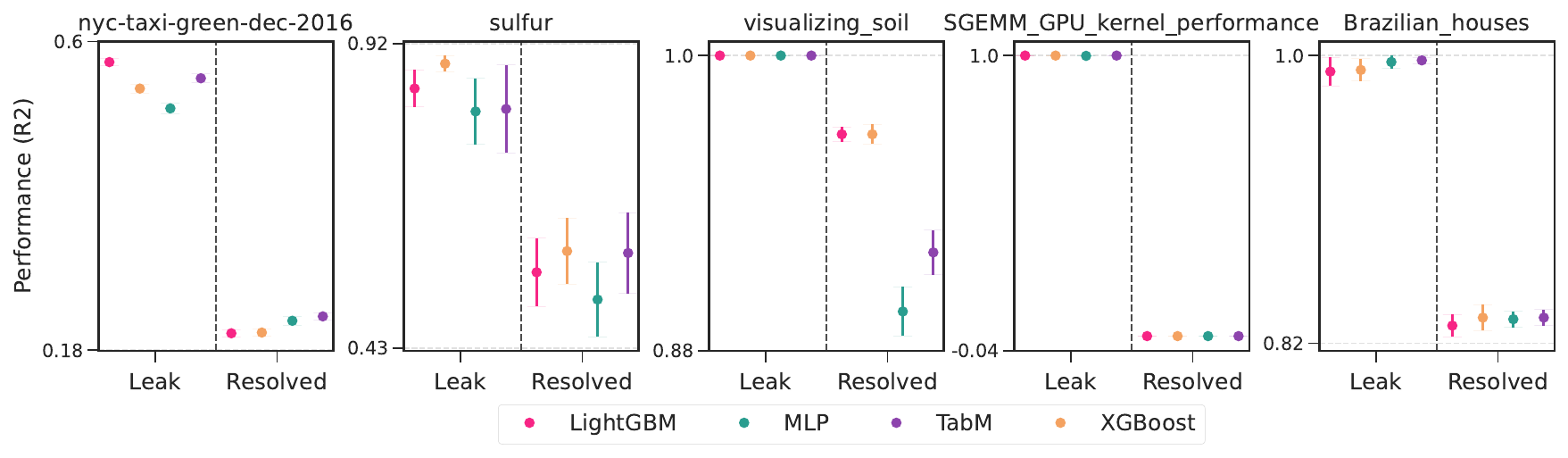}
    \caption{\textbf{Target leaks can alter performance comparisons.} For each dataset, the performance of the models before and after resolving target leaks in the original benchmark is displayed.}
    \label{fig:target_leaks}
\end{figure}

Many of the datasets on OpenML contain features that should not be used as input features for predictive tasks. 
Such features often represent information unavailable during inference in a realistic scenario (e.g., alternative targets), or IDs. 
OpenML provides a functionality that allows specific attributes to be marked as ignored. However, this information is missing or incomplete for many datasets. Thus, users would need to manually exclude these features during preprocessing. This has led to errors when automatically crawling datasets:
\cite{rubachev2024tabred} identified three target leaks in the popular collection of datasets by \cite{grinsztajn2022tree}. Additionally, we identified three more; see Appendix \ref{ssec:reproduce} for a complete list. 
\\
To illustrate the impact of such leaks, we resolve them by removing the leaking features. 
Figure \ref{fig:target_leaks} shows how the performance comparison between different models changes. 
For experimental details and information on dataset-specific impacts, see Appendix \ref{ssec:reproduce}.
For all datasets, target leaks lead to overestimating the performance of all models. 
The effects on the model comparison vary per dataset, but the relative rankings often change. 
Consequently, target leaks can impact the conclusion of a study.



\begin{figure}[t]
    \centering
    \includegraphics[width=0.99\columnwidth]{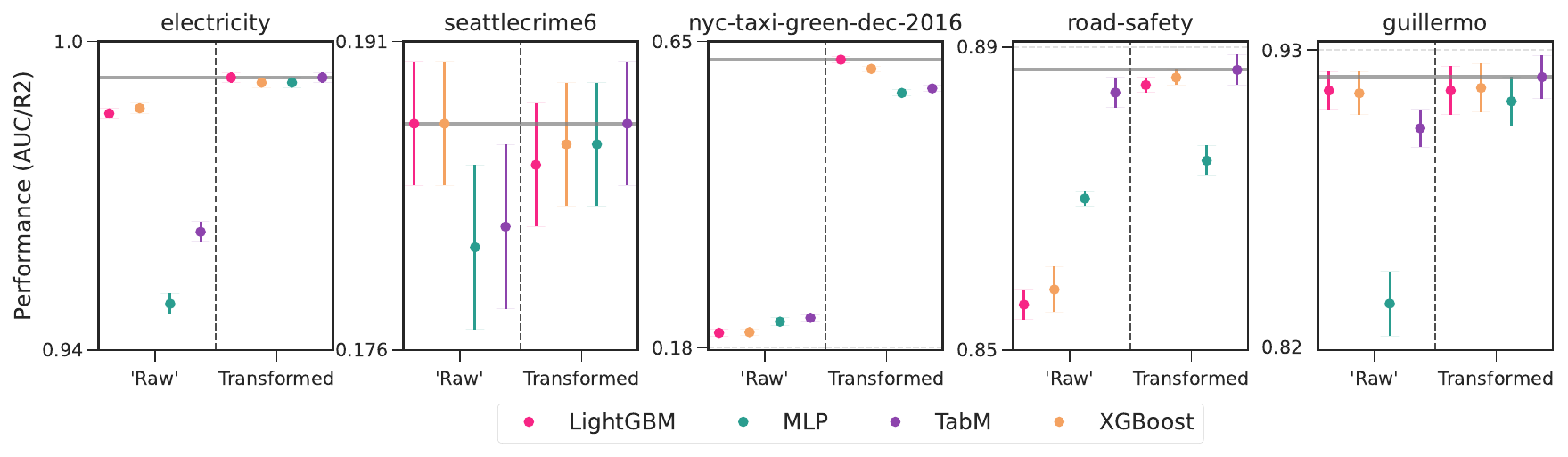}
    \caption{\textbf{Simple yet effective preprocessing decisions can entirely change model comparisons.} AUC performance is displayed for classification tasks, and R2 for regression (seattlecrime \& nyc-taxi). 'Raw' denotes the (already preprocessed) dataset version provided by the benchmark. Transformed denotes a dataset after simple preprocessing. This includes treating ordinal features as categorical (electricity \& seattle), adding the difference between two dates as a feature (nyc-taxi), adding fractions of features as new features (road-safety), and feature selection (guillermo). A horizontal line marks the highest achieved performance.}
    \label{fig:preprocessing_example}
\end{figure}

OpenML currently uses a versioning system where a newer version corresponds to a more recent upload of a dataset with the same name. This misleading system led \citet{grinsztajn2022tree} to use preprocessed instead of original dataset versions for several datasets (e.g., seattlecrime6).
To further complicate this issue, OpenML provides only limited information about the processing history of datasets.
There is no distinction between raw and preprocessed datasets, dataset descriptions rarely provide clarity, and uploaders cannot separately specify preprocessing requirements for datasets.
As a result, users often overlook whether preprocessing is required, apply inadequate preprocessing, or rely on suboptimal preprocessing enforced by the dataset uploader.
Figure \ref{fig:preprocessing_example} illustrates that several datasets greatly benefit from different, preprocessing choices; see Appendix \ref{ssec:reproduce} for details.
These preprocessing choices would likely be among the first implemented by practitioners after inspecting the data.  
Yet, there is no option for such knowledge to be incorporated in OpenML, besides the dataset description.
The preprocessing often benefits all models while drastically impacting the conclusion of model comparisons. In line with \cite{tschalzev2024data}, it can be observed that large performance differences often diminish when applying the appropriate task-specific preprocessing.

\textbf{How can researchers avoid this issue?}  \hspace{1.5mm}
Users of benchmarks should reflect upon which dataset to use and how to use them (e.g., as done by \cite{gorishniy2024tabm}). 
To avoid target leaks, we suggest investigating datasets where the performance of all models is close to perfect on intuitive metrics such as AUC for classification or R2 for regression. 
Furthermore, researchers should always find out how to treat features appropriately for a task. 
In particular, date, categorical, or ordinal features require attention.  
This includes passing the right feature types to the algorithms (e.g., for XGBoost's categorical features handling). 
Many tabular datasets are from real-world applications with semantic meaningful names. 
Thus, if datasets are carefully inspected and their quality is not taken for granted, mistakes can often be avoided.

\textbf{How can data repositories prevent this issue?}   \hspace{1.5mm}
Our examples have clearly shown that even knowledgeable researchers still oversee simple, in hindsight, obvious data issues.
Hence, the community would greatly benefit from data repositories leading action to prevent false usage of datasets.
Regarding misuse of features, the ignore functionality might not suffice. An alternative could be to adapt the dataset schema to provide non-predictive features separately from input and output features.
Instead of connecting datasets by 'versions' when they share a name, we suggest that datasets should be hosted as a combination of raw datasets and preprocessed versions referencing those. 
Furthermore, tasks should always use preprocessed dataset versions and the schema of those should include a separate description of preprocessing steps applied to obtain them from the raw datasets. 
The community could contribute by cleaning up existing datasets.
Having examples of preprocessing pipelines would also provide a basis for evaluating those in studies. Research in this direction is minimal despite playing a pivotal role in real-world tasks \citep{tunguz2023kaggle}.

\section{Towards Data Repositories That Facilitate Appropriate Dataset Use}
\label{sec:potential}
In this section, we present our ideal vision of how OpenML could evolve into an optimal data repository for a researcher focused on advancing the state-of-the-art in predictive machine learning for tabular data.
We emphasize that although our discussion focuses on data repositories, they are not ultimately responsible for issues in research papers.
Nevertheless, following \cite{longjohnbenchmark}, data repositories can serve as an essential intermediary between dataset creators and users. 
Therefore, we propose implementing top-down solutions, led by data repositories, to address the highlighted issues.


\textbf{Suggested improvement.}   \hspace{1.5mm}
Currently, each OpenML task consists of a dataset and an evaluation method for obtaining test scores.
While each task is associated with a dataset, the only evaluation-related information in the dataset schema is the target feature and which features to ignore. 
Our examples have shown that even peer-reviewed studies that received much attention made suboptimal decisions regarding the remaining choices for the evaluation design, preventing accurate claims about the state-of-the-art. 
To mitigate similar issues, we suggest that, in extension to the aspects proposed so far,
\textbf{OpenML should associate raw datasets with recommended default tasks}.
Based on our insights, we argue that these default tasks should ideally contain the following minimum requirements: 1) target, 2) preprocessing protocol, 3) estimation protocol (train/test splits) 4) validation protocol for model selection in HPO, 5) default metric, 6) post-processing requirements, 7) the performance of a strong baseline as a reference point.
Importantly, we do not propose a fixed evaluation setup that all researchers must follow. Instead, we propose providing a reference that remains open for discussion, allowing users to critique it and enabling improvements in future revisions.
Furthermore, users can still define alternative tasks tailored to their specific criteria, and research questions.

\textbf{Benefits.} \hspace{1.5mm} 
A default task would have several major benefits: 
1) beginners will have examples to learn from, fulfilling the educational responsibility of a data repository; 
2) benchmark curators have a starting point for each dataset and can save effort; 
3) users of the datasets for academic purposes would make fewer mistakes; 
4) there would be a relevant baseline to beat for each dataset, s.t. improvements over the state-of-the-art can be made explicit instead of going under in standardized evaluations; 
5) there would be a reference point to improve upon if issues like data leaks are found; 
6) dataset versioning will be more intuitive as it is always w.r.t. the raw dataset and preprocessed versions could be defined in separate tasks. 


\textbf{Who should define default tasks?} \hspace{1.5mm} 
As OpenML is an open-upload community, performing quality checks and defining appropriate default tasks for each available dataset would be tedious. 
Therefore, we suggest starting with the datasets and evaluation setups of an established, carefully curated benchmark such as the AutoML benchmark \cite{gijsbers2024amlb}. 
Once a reference is available, insights from research papers such as \cite{rubachev2024tabred}, \cite{tschalzev2024data}, or this paper can be incorporated to adapt the default tasks. 
For additional validation, we propose highlighting datasets associated with an approved default task using a 'verified task' tag in addition to the currently existing 'verified' tag.
To minimize the workload for repository maintainers, we suggest distributing the responsibility of tracking the performance of strong baselines on default evaluation tasks across the community. 

\section{Related Work}
Our paper builds on the discussion of data and evaluation quality issues by \citet{tschalzev2024data} and \citet{rubachev2024tabred}, contributing to a greater awareness of data repository misuse in tabular data research. While the role of data repositories in benchmarking was recently examined from a general perspective by \citet{longjohnbenchmark}, our study focuses on three key aspects of supervised learning for tabular data.

The importance of model selection strategies has long been known in the general ML literature \citep{cawley2010over}. However, it has not been discussed in modern tabular data research until recently \citep{nagler2024reshuffling}. We contribute to raising awareness of this issue by demonstrating how holdout model selection underestimates model performance on the TabZilla benchmark \citep{mcelfresh2023neural}.
\\
\citet{tschalzev2024data} emphasized that existing studies lack external validity in the absence of a clear definition of strong performance for a given task. However, their analysis focuses exclusively on data from Kaggle competitions, without examining the implications of their findings for existing benchmarks. We validate their conclusions by demonstrating how adding a single strong baseline on the TabZilla benchmark could have mitigated false conclusions regarding model performance differences and prevented the establishment of a misleading baseline for subsequent studies.
\\
\citet{rubachev2024tabred} highlighted that many datasets are ill-suited for the tabular domain (e.g., tables of images) or require custom data splits. Furthermore, the authors found seven datasets with leak issues in the Grinsztajn benchmark~\citep{grinsztajn2022tree}. We extend the findings of \citet{rubachev2024tabred} by investigating the impact of target leaks on model comparisons, resolving them where possible, identifying three more datasets with leak issues, and identifying five datasets where conclusions were contaminated by inappropriate preprocessing. The latter additionally supports the finding of \cite{tschalzev2024data} that the input data often has a more important role than the choice of the model class. 

Overall, our paper follows a line of work critically reflecting evaluation practices in academia \citep{sculley2018winner, zhang2021importance, herrmann2024position}.
Furthermore, we follow the recommendation of \cite{orr2024building} that dataset creators should restrict inappropriate applications. Accordingly, we provide recommendations for data repositories in Section \ref{sec:potential}.

\section{Limitations and Conclusions}


\textbf{Limitations} \hspace{1.5mm} 
\textbf{(1)} We focus on supervised i.i.d. learning and tabular data. Other modalities and tasks may have different requirements possibly colliding with some of our suggestions. 
\textbf{(2)} We focus on state-of-the-art predictive performance, as improving on it is one of the most frequent goals of recent studies. However, academic studies often have different scopes, which is not reflected in our discussion. 
\textbf{(3)} We mainly focus on OpenML as it is the currently most frequently used data repository for tabular data research. 
\textbf{(4)} Data repositories such as OpenML are complex, and we analyzed them to the best of our understanding, yet we do not know the full history of and intentions behind design choices. 
\textbf{(5)} We only focus on three issues as there was no room for more. We certainly did not cover every reason for reduced research quality.
\textbf{(6)} Lastly, we do not offer off-the-shelf solutions to demonstrated issues, as solving them is a structural problem. Nevertheless, by highlighting issues and proposing possible improvements, we contribute an important step towards improved ML evaluation quality.

\textbf{Conclusion} \hspace{1.5mm}
In this paper, we have highlighted the overlooked need for dataset-specific model selection strategies in validation, appropriate strong baselines, and task-specific preprocessing guidelines. We have demonstrated that prior studies were compromised by the unreflected use of data repositories, leading to less reliable conclusions.
By illustrating the impact of questionable practices, we aim to help researchers avoid—and data repositories prevent—pitfalls in future work. To address these issues at a larger scale, we propose the introduction of expert-reviewed default evaluation tasks for datasets, supported by data repositories.
Our work underscores the critical role of data repositories in shaping evaluation quality standards. 


\newpage

\subsubsection*{Acknowledgments}
This research was supported by the Ministry of Economic Affairs, Labour and Tourism Baden-Württemberg. Furthermore, the authors acknowledge support by the state of Baden-Württemberg through bwHPC and the German Research Foundation (DFG) through grant INST 35/1597-1 FUGG.
We acknowledge funding by the Deutsche Forschungsgemeinschaft (DFG,
German Research Foundation) under SFB 1597 (SmallData), grant number 499552394. 
Frank Hutter acknowledges the financial support of the Hector Foundation.


\bibliography{iclr2025_conference}
\bibliographystyle{iclr2025_conference}

\appendix
\section{Experimental Details} \label{sec:experimental_details}

\subsection{Software and Hardware}
The deep learning models and XGBoost were trained using one or more of the following GPU hardware, depending on the availability: NVIDIA H100, NVIDIA RTX A6000, NVIDIA A5000, or NVIDIA A40. LightGBM was trained using the following CPU hardware: Intel(R) Xeon(R) CPU E2640v2 @ 2,00 GHz.

\subsection{Datasets and Preprocessing}

\textbf{TabZilla} \hspace{1.5mm} We use the 36 datasets defined by \cite{mcelfresh2023neural} as the TabZilla-hard benchmark. All datasets are classification tasks with varying no. of classes. For each dataset, we use the train/validation/test splits defined by the benchmark. Note that the train/test splits have been adapted from previous work \citep{bischl2017openml} by the benchmark creators. The benchmark uses the datasets without any preprocessing, unless the one required for specific models. 
In addition, we combine classes that occur less than 3 times in the raw dataset into one for the lymph and audiology datasets, as it is impossible to meaningfully evaluate them. The benchmark authors did not further describe whether and how they deal with this issue. 
\citet{rubachev2024tabred} found three datasets with leak issues in the benchmark: artificial-characters, GesturePhaseSegmentationProcessed, and electricity. We decided to leave the datasets in the benchmark to be able to directly compare to the results obtained by \cite{mcelfresh2023neural}. 
Nevertheless, for the first two of these datasets, there is no clear procedure to resolve the leaks in the random split setting. Therefore, following \citet{rubachev2024tabred}, we recommend not to use these datasets with random splits. 
For the electricity dataset, random splits can still correspond to meaningful tasks. All of the models we used can exploit the information from the 'leak' equally well when treating the date feature as categorical (see Figure \ref{fig:preprocessing_example}). Hence, this dataset could still be used, although due to all models achieving strong performance, the task may be too trivial, depending on the purpose of the evaluation.

\textbf{\cite{grinsztajn2022tree} Benchmark} \hspace{1.5mm}
The original paper introduced a benchmark where datasets where evaluated divided in four subtasks by binary classification vs. regression and by whether categorical features are in the dataset. However, in subsequent studies, this evaluation setup was often adapted s.t. the 45 unique datasets were combined into one evaluation \citep{gorishniy2023tabr,holzmuller2024better}. We use the benchmark datasets as follows:
\begin{enumerate}
    \item Whenever there are multiple versions of a dataset, we use the version closest to the unpreprocessed version to the dataset: (1) When a task is available as regression and classification, we use the regression task. Whenever a task is available with and without categorical features, we use it with categorical features. Overall, the benchmark consists of 28 regression tasks and 17 binary classification tasks. 
    \item \cite{grinsztajn2022tree} define binary features as categorical, although, binary features are more commonly treated as numeric features. Without binary features, there are only two classification datasets with categorical features remaining (10 for regression). Moreover, only 5 datasets contain categorical features with more than 10 clusters s.t. most categorical features can easily be one-hot-encoded and treated as binary. For these reasons, we do not use this benchmark to evaluate numeric and categorical data separately and also strongly recommend that for follow-up work.
    \item To obtain data splits, we follow the split procedure of \citet{grinsztajn2022tree}: 70\% of the available data is used for training. Of the remaining 
    30\%, 30\% are used for validation and 70\% for testing. Train sets are truncated to 10,000 samples, and validation and test sets are truncated to 50,000, as in the original benchmark. Our evaluation differs from the original benchmark in two aspects: (1) The original benchmark used 1-5 split repetitions (folds) depending on the test size. In contrast, we use 15 repetitions for all datasets to obtain a more robust evaluation quantifying the involved randomness. (2) \citet{grinsztajn2022tree} separate an additional validation set for early stopping from the train data. In contrast, we use the same as is used for hyperparameter selection to better utilize the available data without exceeding the original benchmark logic of training on medium sized datasets ($\leq10,000$ train samples). 
\end{enumerate}

\textbf{Model-Specific Preprocessing} \hspace{1.5mm}
For the tree-based models, model-specific preprocessing only included the correct assignment of datatypes to categorical features. For the deep learning models, the preprocessing was defined in line with the related work \citep{gorishniy2021revisiting,grinsztajn2022tree}. For regression, the target is normalized to zero mean and unit variance. For numeric features, missing values are replaced with the mean, and the features are normalized using ScikitLearn’s QuantileTransformer \citep{scikit-learn}. For MLPs, categorical features are ordinally encoded and processed by entity embeddings \citep{guo2016entity}. For TabM, we use one-hot-encoding following the author's recommendations. 

\subsection{Model Training and Hyperparameter Optimization} 
We use the optuna library \citep{akiba2019optuna} for hyperparameter optimization. Each model is optimized for 100 trials with the first 20 trials being random search trials and the remaining 80 using the multivariate Tree-structured Parzen Estimator algorithm \citep{bergstra2011algorithms,falkner2018bohb}.
The models are trained and tuned using cross entropy for classification and mean squared error for regression. The AdamW optimizer is used for training the deep learning models \citep{loshchilov2017decoupled}.
We use logloss and mean squared error for early stopping and validation during hyperparameter optimization. 
For XGBoost and LightGBM we use the search space defined by \cite{tschalzev2024data}, which itself is mostly taken from other related work \citep{gorishniy2021revisiting,grinsztajn2022tree} and the library documentations. 
For MLPs and TabM, we use the search spaces provided by the authors \citep{gorishniy2021revisiting, gorishniy2024tabm} -- with some adjustments as we use categorical feature embeddings for MLPs.
We evaluate each of the included models with an equal number of hyperparameter trials. Therefore, we did not use time budgets to constrain the number of trials per model and dataset. As this leads to long computation times for some models, we restricted the optimization to 20 random search trials for the following datasets in the TabZilla benchmark: guillermo ($>4000$ features), albert (many high-cardinality categorical features), airlines (large sample size), and poker-hands (large sample size). This HPO regime still exceeds the evaluation of \cite{mcelfresh2023neural}, since many models failed for these datasets entirely.
All search spaces are provided in Tables \ref{tab:hp-xgb}-\ref{tab:hp-tabm}.

\begin{table}[h]
    \centering
    \small
    \begin{tabular}{lll}
    \toprule
    Hyperparameter & Default &   Search distribution \\
    \midrule
    n\_estimators & 4000 & - \\
    patience & 200 & - \\
    learning\_rate & 0.3 & LogUniform[1e-3, 0.7] \\
    max\_depth & 6 & UniformInt[1, 11] \\
    colsample\_bytree & 1. & Uniform[0.5,1.] \\
    subsample & 1. &  Uniform[0.5,1.]\\
    min\_child\_weight & 1. & LogUniform[1, 100] \\
    reg\_alpha & 0. & LogUniform[1e-8, 100] \\
    reg\_lambda & 1. & LogUniform[1, 4] \\
    gamma & 0. & LogUniform[1e-8, 7] \\
    
    \bottomrule
    \end{tabular}
    \caption{Hyperparameter configurations for XGBoost.}
    \label{tab:hp-xgb}
\end{table}

\begin{table}[h]
    \centering
    \small
    \begin{tabular}{lll}
    \toprule
    Hyperparameter & Default &   Search distribution \\
    \midrule
    iterations & 4000 & - \\
    patience & 200 & - \\
    learning\_rate & 0.1 & LogUniform[1e-3, 0.7] \\
    max\_depth & -1 & \{-1, UniformInt[1, 11]\} \\
    min\_data\_in\_leaf & 20 & \{20, 50, 100, 500, 1000, 2000\} \\
    num\_leaves & 31 & UniformInt[2, 2047] \\
    lambda\_l2 & 0. & LogUniform[1e-4, 10.] \\
    feature\_fraction & 1. &  Uniform[0.5, 1.]  \\
    bagging\_fraction & 1. &  Uniform[0.5, 1.]  \\
    min\_sum\_hessian\_in\_leaf & 1e-3 & LogUniform[1e-4,100.0] \\
     
    \bottomrule
    \end{tabular}
    \caption{Hyperparameter configurations for LightGBM. If max\_depth$\geq 1$, the possible num\_leaves ranges are adjusted to be in a space feasible with the respective depth.}
    \label{tab:hp-lgb}
\end{table}

\begin{table}[h]
    \centering
    \small
    \begin{tabular}{lll}
    \toprule
    Hyperparameter & Default &   Search distribution \\
    \midrule
    epochs & 200 & - \\
    patience & 5 & - \\
    batch\_size & 128 & - \\
    learning\_rate & 1e-3 & LogUniform[1e-5, 1e-2] \\
    weight\_decay & 1e-4 & LogUniform[1e-6, 1e-3] \\
    \# Layers &  2 & UniformInt[1, 6] \\ 
    Layer size & 128 & UniformInt[64, 1024] \\ 
    Dropout & 0.25 & Uniform[0., 0.5] \\
    Categorical embedding size & 8 & UniformInt[1, 512] \\       \bottomrule
    \end{tabular}
    \caption{Hyperparameter configurations for MLPs \citep{gorishniy2021revisiting}.}
    \label{tab:hp-mlpp}
\end{table}

\begin{table}[h]
    \centering
    \small
    \begin{tabular}{lll}
    \toprule
    Hyperparameter & Default &   Search distribution \\
    \midrule
    epochs & 200 & - \\
    patience & 5 & - \\
    batch\_size & 128 & - \\
    k (\# of ensemble adapters \& heads) & 32 & 32 \\
    learning\_rate & 0.002 & LogUniform[1e-4, 5e-3] \\
    weight\_decay & 0.0003 & \{0, LogUniform[1e-4, 1e-1]\} \\
    \# Layers &  3 & UniformInt[1, 5] \\ 
    Layer size & 512 & UniformInt[64, 1024] \\ 
    Dropout & 0.1 & \{0, Uniform[0., 0.5]\} \\
    Categorical embedding size & 8 & UniformInt[1, 128] \\ 
    Numerical embedding size & 8 & UniformInt[1, 128] \\ 
    \bottomrule
    \end{tabular}
    \caption{Hyperparameter configurations for TabM \cite{gorishniy2024tabm}.}
    \label{tab:hp-tabm}
\end{table}

\subsection{Metrics and Evaluation} \label{app:metrics}
All models are trained to optimize logloss for classification and mean squared error for regression. Hyperparameter optimization is also conducted to tune the models for these metrics, and after HPO, the best model is selected according to these metrics. Therefore, the main metrics we use throughout the paper are logloss and RMSE. For Figures \ref{fig:target_leaks} and \ref{fig:preprocessing_example}, we use AUC and R2 to better show that some models reach close to perfect performance, which is more intuitive to understand with those metrics. However, the reported tendencies and conclusions are the same for both types of metrics. Whenever reporting results on single datasets, we choose the best model according to logloss/RMSE on validation data. Whenever aggregating over datasets, we follow the procedure of \citet{mcelfresh2023neural}: The normalized average logloss (Table \ref{tab:validity}), is computed per fold. We use the average distance to the minimum (ADTM) metric to transform each fold to be in $[0,1]$. In general, using this metric has flaws like overemphasizing small but irrelevant differences, and we would not recommend relying solely on this metric. However, we report it to compare to the evaluation of \cite{mcelfresh2023neural} and, in addition, report average ranks and unnormalized metrics. Ranks were computed by first computing ranks in each fold before aggregation.

\newpage
\subsection{Additional Details for Tables and Figures} \label{ssec:reproduce}
In addition to the information provided in the captions and main text, we provide additional context for the reported figures and tables:
\begin{enumerate}
    \item For \textbf{Figure} \ref{fig:5cv_gains}, five datasets from the TabZilla benchmark were not used: \textbf{(1)} For the monks-problems-2 dataset, all models were able to perfectly generalize on the test data. This is another issue since the dataset was included in the benchmark as it was supposed to be 'hard for GBDTs'. \textbf{(2)} The guillermo, albert, airlines, and poker-hands datasets were not used as 5-fold cross-validation for model selection was infeasible due to the long training time required. However, in real-world applications, practitioners would likely avoid holdout validation, and instead preprocess the dataset or adjust model search spaces to enable advanced model selection techniques. As we wanted the models and datasets to stay consistent with the original benchmark, we decided to leave these datasets out. 
    \item The examples in \textbf{Figure} \ref{fig:four_examples_valoverfitting} were selected to provide tangible examples of the impact of different model selection techniques. While these examples were hand-picked to fit our narrative, we acknowledge that other examples could have told a different story. Nevertheless, as Figure \ref{fig:5cv_gains} indicates, only a few examples contradict our findings, whereas the overall trend strongly supports our claims.
    \item For comparing to the original TabZilla benchmark, we used the results uploaded by \cite{mcelfresh2023neural}, which they referenced in their Git repository\footnote{\url{https://github.com/naszilla/tabzilla/blob/main/TabZilla/tutorials/analyze_metadataset.ipynb}}. For \textbf{Table} \ref{tab:validity}, we use the provided full HPO search results to select the best model according to logloss for all models trained by \citet{mcelfresh2023neural}. We concatenate the results of our models and calculate the reported aggregated metrics using all performance results.
    \item For \textbf{Table} \ref{tab:corr}, we use the results reported by \cite{mcelfresh2023neural} in their Table 14. Furthermore, we use their uploaded results to conduct the same analysis, but reducing the no. of datasets to the 36 datasets chosen for the benchmark. First, we reproduce the analysis solely with their uploaded results, and second, we replaced their versions of LightGBM \& MLPs with our improved HPO setups: 100 trials with holdout validation and 100 trials with 5-fold cross-validation.  
    \item Details on the leaks \footnote{The leaks for the sulfur, visualizing\_soil, and SGEMM dataset were found by \cite{rubachev2024tabred}. Furthermore, they discovered the following leak issues: grouped data leaks in the eye\_movements dataset, temporal leaks in the electricity and Bike\_sharing\_demand datasets, and duplicate data leakage in the compass dataset.} in the \citet{grinsztajn2022tree} benchmark and how we resolved them in Figure \ref{fig:target_leaks}:
    \begin{enumerate}
        \item \textit{nyc-taxi-green-dec-2016}: The goal is to predict the tip amount given by a taxi passenger. However, since the total amount—of which the tip is a component—is included, this results in target leakage. In Figure \ref{fig:target_leaks}, we resolve the leak by excluding the total amount, as suggested in the OpenML description for the dataset\footnote{\url{https://www.openml.org/d/42729}}. 
                
        \item \textit{sulfur}:  Two related targets exist, one of which was mistakenly used as an input feature. In Figure \ref{fig:target_leaks}, we resolve the issue by removing the erroneous input feature.  
                
        \item \textit{visualizing\_soil}: The benchmark uses the wrong target, making the task trivial. In Figure \ref{fig:target_leaks}, we try to resolve the leak by using the original target feature and treating the wrongly used target as a categorical input. However, the performance of tree-based models over neural networks is unusually large. Therefore, we hypothesize that there are other issues remaining with this dataset.  
               
        \item \textit{SGEMM\_GPU\_kernel\_performance}: The dataset has four target features, three of which were mistakenly included as inputs in the benchmark. In Figure \ref{fig:target_leaks}, we resolve the issue by removing the alternative targets from the input space. However, the remaining features defined by the benchmark do not allow to learn anything better than a random baseline. Hence, the dataset should not be used for predictive modeling. 
                
        \item \textit{Brazilian\_houses}:  The task is to predict the total house value, but rent\_amount, property\_tax, and fire\_insurance are part of the inputs making the task trivial. Therefore, in Figure \ref{fig:target_leaks}, we exclude these features. 
        
        \item \textit{medical\_charges}: The task is to predict the average total payments of hospital patients at discharge. After the preprocessing of \citet{grinsztajn2022tree}, the only remaining features are quantities which are part of the target (Total\_Discharges, Average\_Covered\_Charges, Average\_Medicare\_Payments). Since no features are left after removing those, this dataset cannot be used anymore. \\

    \end{enumerate}
    \item Details on the transformations applied to the datasets in Figure \ref{fig:preprocessing_example}:
    \begin{enumerate}
        \item electricity: \cite{rubachev2024tabred} found that samples collected closer in time are related for this dataset, leading to near-perfect predictions for retrieval-based models. However, we argue that using random splits is not necessarily forbidden for such datasets. The dataset, as it is used in current benchmarks, can also be seen as a task, where the goal is to fill missing target values in a time-series. With this task conceptualization the main issue of current benchmarks is that the date feature is encoded as ordinal and treated as numeric, favoring tree-based models like XGBoost over neural networks. In Figure \ref{fig:preprocessing_example}, we address this by treating the date feature as categorical, which benefits all models and enables them to achieve similarly high performance, close to a perfect AUC.
        
        \item \textit{seattlecrime6}: The benchmark uses a preprocessed version of the dataset where the task is to predict a 'Reported time' feature. However a description what this feature means is missing and it greatly differs from the original feature in the raw dataset. While the alternative target is not a technical issue, using this dataset version does not correspond to a meaningful real-world task anymore. A technical issue, however, is that time features (hour and minute) were preprocessed using ordinal encoding and treated as numeric in the benchmark. In Figure \ref{fig:preprocessing_example}, we resolve this by treating these features as categorical. We suspect that OpenML's misleading versioning system led to the use of this suboptimally preprocessed dataset version.
        
        \item \textit{nyc-taxi-green-dec-2016}: Even after resolving the target leak, the dataset remains problematic. The benchmark uses a dataset version where the start time and end time features were already preprocessed into day, hour, and minute. However, a more informative feature would be the difference between the two original datetime values. Applying this transformation reverses the performance tendencies of tree-based models versus neural networks (Figure \ref{fig:preprocessing_example}).
        
        \item \textit{road-safety}: The goal is to predict the sex\footnote{
        Please note that we consider this prediction task to be ethically dubious. Similar to other cases like this before (\url{https://github.com/scikit-learn/scikit-learn/issues/16155}), we believe the use of \textit{road-safety} should be discontinued and not be part of a default benchmark for the community.} of the driver in road accidents. To enhance model performance, as shown in Figure \ref{fig:preprocessing_example}, we compute all possible ratios between numeric features. This improves the performance of all models, particularly tree-based ones.
        
        \item \textit{guillermo}: The dataset contains over 4,000 features. For practical reasons, data scientists aiming for high performance on this dataset would likely begin by removing redundant or irrelevant features. Therefore, using all available features may not always be the best approach, depending on the research question. In Figure \ref{fig:preprocessing_example}, we select the 200 most important features based on feature importance scores from a default LightGBM model. As shown, this procedure benefits all models, including LightGBM itself. Notably, after feature selection, TabM slightly outperforms tree-based models on average, whereas the latter were clearly superior before. This further supports the findings of \cite{grinsztajn2022tree}, which suggest that neural networks are more negatively impacted by irrelevant features than gradient-boosted trees.
    \end{enumerate}
\end{enumerate}



\section{Full Results} \label{sec:full_results}

\begin{longtable}{lllll}
    \toprule
     & lymph & audiology & heart-h & colic \\
    \midrule
    LightGBM (Holdout) & \textbf{0.4926 (0.2858)} & 1.1609 (0.247) & 0.4586 (0.0956) & \textbf{0.4427 (0.1196)} \\
XGBoost (Holdout) & \textbf{0.452 (0.2003)} & 0.8044 (0.282) & 0.4485 (0.0645) & \textbf{0.3685 (0.0863)} \\
MLP (Holdout) & \textbf{0.7361 (0.3783)} & 0.9935 (0.7361) & \textbf{0.523 (0.3021)} & \textbf{0.5986 (0.4158)} \\
TabM (Holdout) & \textbf{0.4517 (0.2791)} & 0.5583 (0.1653) & \textbf{0.4232 (0.1026)} & \textbf{0.3698 (0.1052)} \\
LightGBM (5CV) & \textbf{0.449 (0.187)} & 1.04 (0.127) & \textbf{0.408 (0.057)} & 0.387 (0.061) \\
XGBoost (5CV) & \textbf{0.35 (0.132)} & 0.672 (0.2) & \textbf{0.416 (0.058)} & \textbf{0.333 (0.065)} \\
MLP (5CV) & \textbf{0.352 (0.129)} & \textbf{0.469 (0.155)} & \textbf{0.376 (0.089)} & \textbf{0.381 (0.08)} \\
TabM (5CV) & \textbf{0.327 (0.123)} & \textbf{0.467 (0.137)} & \textbf{0.376 (0.071)} & \textbf{0.34 (0.081)} \\
    \bottomrule
    
    \toprule
     & monks-problems-2 & balance-scale & profb & Australian \\
    \midrule
LightGBM (Holdout) & \textbf{0.0 (0.0)} & \textbf{0.0607 (0.0741)} & \textbf{0.5692 (0.073)} & \textbf{0.3572 (0.0722)} \\
XGBoost (Holdout) & 0.0152 (0.003) & 0.093 (0.0355) & \textbf{0.6023 (0.0705)} & \textbf{0.3288 (0.0472)} \\
MLP (Holdout) & \textbf{0.0204 (0.0389)} & \textbf{0.0476 (0.0339)} & 0.6131 (0.0783) & 0.3979 (0.0554) \\
TabM (Holdout) & 0.0005 (0.0008) & \textbf{0.0468 (0.0216)} & 0.5944 (0.0623) & \textbf{0.3413 (0.0692)} \\
LightGBM (5CV) & \textbf{-} & \textbf{0.053 (0.027)} & \textbf{0.566 (0.059)} & \textbf{0.305 (0.044)} \\
XGBoost (5CV) & - & 0.07 (0.016) & \textbf{0.553 (0.06)} & \textbf{0.312 (0.045)} \\
MLP (5CV) & \textbf{-} & \textbf{0.037 (0.008)} & \textbf{0.556 (0.045)} & \textbf{0.323 (0.053)} \\
TabM (5CV) & - & \textbf{0.046 (0.014)} & \textbf{0.552 (0.052)} & \textbf{0.328 (0.053)} \\
    \bottomrule
    
    \toprule
     & credit-approval & vehicle & credit-g & qsar-biodeg \\
    \midrule
LightGBM (Holdout) & \textbf{0.3851 (0.2009)} & 0.4915 (0.0539) & 0.533 (0.0469) & \textbf{0.3534 (0.0779)} \\
XGBoost (Holdout) & \textbf{0.3431 (0.1346)} & 0.4587 (0.0367) & \textbf{0.5182 (0.0494)} & \textbf{0.3343 (0.0581)} \\
MLP (Holdout) & \textbf{0.3355 (0.0996)} & 0.4283 (0.0716) & \textbf{0.5128 (0.0364)} & \textbf{0.3508 (0.0829)} \\
TabM (Holdout) & 0.3434 (0.0987) & \textbf{0.3752 (0.0333)} & \textbf{0.504 (0.0342)} & \textbf{0.3121 (0.0766)} \\
LightGBM (5CV) & \textbf{0.307 (0.088)} & 0.453 (0.045) & \textbf{0.499 (0.036)} & \textbf{0.309 (0.059)} \\
XGBoost (5CV) & \textbf{0.309 (0.088)} & 0.437 (0.044) & \textbf{0.489 (0.033)} & \textbf{0.307 (0.06)} \\
MLP (5CV) & \textbf{0.33 (0.092)} & \textbf{0.374 (0.05)} & \textbf{0.491 (0.033)} & \textbf{0.323 (0.072)} \\
TabM (5CV) & \textbf{0.334 (0.077)} & \textbf{0.37 (0.052)} & \textbf{0.49 (0.032)} & \textbf{0.31 (0.063)} \\
    \bottomrule
    
    \toprule
     & cnae-9 & socmob & one-hundred- & mfeat-fourier \\
     &  &  & -plants-texture &  \\
    \midrule
LightGBM (Holdout) & 0.4437 (0.0848) & \textbf{0.1286 (0.0538)} & 0.7639 (0.1135) & 0.4215 (0.0517) \\
XGBoost (Holdout) & 0.2367 (0.0671) & \textbf{0.114 (0.0325)} & 0.8602 (0.0755) & 0.4247 (0.0453) \\
MLP (Holdout) & 0.176 (0.0594) & 0.1724 (0.0595) & 0.498 (0.0717) & 0.4765 (0.075) \\
TabM (Holdout) & 0.1525 (0.0656) & 0.1288 (0.0403) & \textbf{0.4523 (0.0639)} & \textbf{0.3426 (0.0523)} \\
LightGBM (5CV) & 0.428 (0.064) & \textbf{0.116 (0.035)} & 0.637 (0.081) & 0.39 (0.045) \\
XGBoost (5CV) & 0.236 (0.064) & \textbf{0.106 (0.027)} & 0.834 (0.075) & 0.398 (0.043) \\
MLP (5CV) & \textbf{0.118 (0.04)} & \textbf{0.116 (0.027)} & \textbf{0.427 (0.074)} & 0.426 (0.049) \\
TabM (5CV) & \textbf{0.136 (0.053)} & \textbf{0.108 (0.029)} & \textbf{0.421 (0.066)} & \textbf{0.329 (0.036)} \\
    \bottomrule
    
    \toprule
     & mfeat-zernike & kc1 & jasmine & splice \\
    \midrule
LightGBM (Holdout) & 0.4879 (0.0599) & 0.3565 (0.0325) & 0.4107 (0.0293) & 0.1162 (0.0244) \\
XGBoost (Holdout) & 0.4865 (0.0592) & 0.3483 (0.0247) & 0.4004 (0.0195) & \textbf{0.1003 (0.0175)} \\
MLP (Holdout) & 0.3801 (0.0882) & \textbf{0.344 (0.0202)} & 0.4339 (0.034) & 0.1385 (0.0383) \\
TabM (Holdout) & \textbf{0.2994 (0.0423)} & \textbf{0.342 (0.0258)} & 0.4085 (0.0244) & 0.1155 (0.0212) \\
LightGBM (5CV) & 0.468 (0.058) & \textbf{0.34 (0.022)} & \textbf{0.395 (0.025)} & 0.11 (0.019) \\
XGBoost (5CV) & 0.461 (0.055) & \textbf{0.339 (0.022)} & \textbf{0.391 (0.023)} & \textbf{0.096 (0.016)} \\
MLP (5CV) & \textbf{0.322 (0.06)} & \textbf{0.338 (0.021)} & 0.427 (0.025) & 0.122 (0.026) \\
TabM (5CV) & \textbf{0.301 (0.04)} & \textbf{0.335 (0.019)} & 0.403 (0.022) & 0.11 (0.022) \\
    \bottomrule
    
    \toprule
     & Bioresponse & ada\_agnostic & phoneme & SpeedDating \\
    \midrule
LightGBM (Holdout) & \textbf{0.4377 (0.0283)} & \textbf{0.323 (0.0272)} & \textbf{0.2353 (0.029)} & 0.3069 (0.0142) \\
XGBoost (Holdout) & \textbf{0.4382 (0.0318)} & \textbf{0.32 (0.0251)} & \textbf{0.2341 (0.0297)} & 0.3012 (0.0165) \\
MLP (Holdout) & 0.4783 (0.0372) & 0.3418 (0.0225) & 0.2891 (0.0181) & 0.3179 (0.0246) \\
TabM (Holdout) & 0.46 (0.0318) & 0.3416 (0.0207) & \textbf{0.2433 (0.0319)} & 0.3095 (0.0106) \\
LightGBM (5CV) & \textbf{0.431 (0.026)} & \textbf{0.317 (0.021)} & \textbf{0.227 (0.022)} & 0.3 (0.013) \\
XGBoost (5CV) & \textbf{0.429 (0.027)} & \textbf{0.316 (0.022)} & \textbf{0.226 (0.023)} & \textbf{0.296 (0.013)} \\
MLP (5CV) & 0.462 (0.027) & 0.338 (0.022) & 0.279 (0.023) & 0.309 (0.014) \\
TabM (5CV) & 0.449 (0.023) & 0.336 (0.022) & \textbf{0.232 (0.018)} & 0.308 (0.011) \\
    \bottomrule
    
    \toprule
     & GesturePhaseSeg- & elevators & artificial- & nomao \\
     & mentationProcessed &  & characters &  \\
    \midrule
LightGBM (Holdout) & \textbf{0.788 (0.0288)} & 0.2501 (0.0103) & 0.1901 (0.0316) & 0.0721 (0.0069) \\
XGBoost (Holdout) & \textbf{0.7905 (0.0286)} & 0.2436 (0.0099) & 0.2088 (0.028) & 0.0713 (0.0061) \\
MLP (Holdout) & 0.9617 (0.0589) & 0.2826 (0.0192) & 0.6026 (0.0193) & 0.098 (0.0078) \\
TabM (Holdout) & \textbf{0.7726 (0.0378)} & \textbf{0.2427 (0.0105)} & 0.4791 (0.0264) & 0.0866 (0.0059) \\
LightGBM (5CV) & \textbf{0.784 (0.028)} & \textbf{0.242 (0.012)} & \textbf{0.169 (0.021)} & \textbf{0.069 (0.006)} \\
XGBoost (5CV) & \textbf{0.786 (0.029)} & \textbf{0.239 (0.008)} & 0.192 (0.019) & \textbf{0.069 (0.006)} \\
MLP (5CV) & 0.932 (0.036) & 0.288 (0.014) & 0.601 (0.027) & 0.094 (0.005) \\
TabM (5CV) & \textbf{0.781 (0.028)} & \textbf{0.249 (0.012)} & 0.516 (0.024) & 0.086 (0.005) \\
    \bottomrule
    
    \toprule
     & guillermo & jungle\_chess\_- & electricity & higgs \\
     &  & -2pcs-\_raw\_end- &  &  \\
     &  & -game\_complete &  &  \\
    \midrule
LightGBM (Holdout) & \textbf{0.3581 (0.0132)} & 0.2348 (0.004) & 0.149 (0.008) & 0.5236 (0.0083) \\
XGBoost (Holdout) & \textbf{0.3587 (0.0147)} & 0.2392 (0.0054) & 0.1459 (0.0074) & 0.5221 (0.008) \\
MLP (Holdout) & 0.498 (0.02) & 0.0907 (0.0113) & 0.2906 (0.0061) & 0.5291 (0.0059) \\
TabM (Holdout) & 0.384 (0.0116) & \textbf{0.0128 (0.0013)} & 0.2438 (0.0071) & \textbf{0.5072 (0.0081)} \\
LightGBM (5CV) & \textbf{-} & 0.227 (0.005) & 0.144 (0.005) & 0.521 (0.007) \\
XGBoost (5CV) & \textbf{-} & 0.238 (0.005) & \textbf{0.141 (0.005)} & 0.52 (0.008) \\
MLP (5CV) & - & 0.073 (0.007) & 0.28 (0.005) & 0.524 (0.006) \\
TabM (5CV) & - & \textbf{0.012 (0.001)} & 0.244 (0.006) & \textbf{0.506 (0.007)} \\
    \bottomrule

    \toprule
     & MiniBooNE & albert & airlines & poker-hand \\
    \midrule
LightGBM (Holdout) & 0.1291 (0.0051) & 0.5834 (0.0019) & \textbf{0.5995 (0.0013)} & 0.0164 (0.0039) \\
XGBoost (Holdout) & 0.1292 (0.0049) & \textbf{0.5692 (0.0023)} & \textbf{0.5994 (0.0012)} & 0.0118 (0.0007) \\
MLP (Holdout) & 0.1244 (0.0051) & \textbf{0.5705 (0.0021)} & 0.6057 (0.0013) & 0.0004 (0.0002) \\
TabM (Holdout) & 0.1195 (0.0052) & \textbf{0.5692 (0.0028)} & 0.604 (0.001) & \textbf{0.0001 (0.0001)} \\
LightGBM (5CV) & 0.128 (0.005) & - & \textbf{-} & - \\
XGBoost (5CV) & 0.128 (0.005) & \textbf{-} & \textbf{-} & - \\
MLP (5CV) & 0.122 (0.004) & \textbf{-} & - & - \\
TabM (5CV) & \textbf{0.118 (0.005)} & \textbf{-} & - & \textbf{-} \\
    \bottomrule
    \caption{Logloss performance results on the TabZilla-hard benchmark. Bold highlights the best results per column as well as those not significantly different from the best in a Wilcoxon Signed-Rank Test for paired samples with Holm-Bonferroni correction and $\alpha=0.05$. Note that if one method is highlighted in bold and another is not, no conclusions can be drawn between them because the significance tests were conducted only with respect to the best model.}       

    \label{tab:tabzilla_full}

\end{longtable}

\begin{table}[tb]
    \caption{R2 performance results on the Grinsztajn benchmark. Bold highlights the best results per row as well as those not significantly different from the best in a Wilcoxon Signed-Rank Test for paired samples with Holm-Bonferroni correction and $\alpha=0.05$}
    \centering
    \small

    \begin{tabular}{lllll}
    \toprule
    {} & LightGBM & XGBoost & MLP & TabM \\
    \midrule
    nyc-taxi-green-dec-2016 & \textbf{0.5698 (0.0042)} & 0.5343 (0.0068) & 0.5079 (0.0074) & 0.5502 (0.007) \\
    \ \ - Leak resolved & 0.2021 (0.0052) & 0.2037 (0.0049) & 0.2191 (0.0061) & \textbf{0.2257 (0.0056)} \\
    \midrule
    sulfur & 0.8405 (0.0333) & \textbf{0.8837 (0.0129)} & 0.7907 (0.0525) & 0.7945 (0.0825) \\
    \ \ - Leak resolved &\textbf{ 0.5637 (0.0635)} & \textbf{0.5937 (0.0604)} & 0.5044 (0.07) & \textbf{0.5962 (0.0731)} \\
    \midrule
    visualizing\_soil & \textbf{1.0 (0.0)} & 0.9998 (0.0) & 0.9998 (0.0) & 0.9999 (0.0) \\
    \ \ - Leak resolved & \textbf{0.9683 (0.0033)} & \textbf{0.9675 (0.0049)} & 0.8951 (0.0079) & 0.9222 (0.0088) \\
    \midrule
    SGEMM\_GPU\_kernel\_performance & 0.9997 (0.0) & \textbf{0.9997 (0.0)} & 0.9994 (0.0001) & 0.9996 (0.0) \\
    \ \ - Leak resolved & \textbf{0.012 (0.0014)} &\textbf{0.0121 (0.0014)} & \textbf{0.012 (0.0015)} & \textbf{0.0121 (0.0015)} \\
    \midrule
    Brazilian\_houses & 0.9927 (0.0033) & 0.9937 (0.0028) & \textbf{0.9971 (0.0016)} & \textbf{0.9973 (0.0016)} \\
    \ \ - Leak resolved & \textbf{0.8306 (0.0077)} & \textbf{0.8359 (0.006)} & \textbf{0.8327 (0.0042)} & \textbf{0.8346 (0.0042)} \\
    \bottomrule
    \end{tabular}
        \label{tab:target_leaks}
\end{table}

\begin{table}[tb]
    \caption{Performance results on the datasets where we found preprocessing to be helpful. R2 is reported for seattlecrime6 and nyc-taxi-green-dec-2016, AUC for the remaining datasets. Bold highlights the best results per column as well as those not significantly different from the best in a Wilcoxon Signed-Rank Test for paired samples with Holm-Bonferroni correction and $\alpha=0.05$. Note that if one method is highlighted in bold and another is not, no conclusions can be drawn between them because the significance tests were conducted only with respect to the best model. Additionally, note that electricity and guillermo are from the TabZilla benchmark, while the other datasets are from the Grinsztajn benchmark.}
    \centering
    \small

    \begin{tabular}{llllll}
    \toprule
     & electricity & seattlecrime6 & nyc-taxi-green-dec-2016 & road-safety & guillermo \\
    \midrule
    LightGBM & 0.986 (0.001) & \textbf{0.186 (0.003)} & 0.202 (0.005) & 0.856 (0.002) & 0.915 (0.007) \\
     + preprocessing & \textbf{0.993 (0.001)} & 0.185 (0.003) & \textbf{0.621 (0.005)} & 0.885 (0.001) & 0.915 (0.01) \\
    XGBoost & \textbf{0.987 (0.001)} & \textbf{0.186 (0.003)} & 0.204 (0.005) & 0.858 (0.002) & 0.914 (0.008) \\
     + preprocessing & 0.992 (0.001) & \textbf{0.186 (0.004)} & 0.608 (0.005) & 0.886 (0.001) & 0.916 (0.009) \\
    MLP & 0.949 (0.002) & 0.18 (0.005) & 0.219 (0.006) & 0.87 (0.002) & 0.836 (0.013) \\
     + preprocessing & 0.992 (0.001) & 0.185 (0.004) & 0.571 (0.006) & 0.874 (0.002) & 0.911 (0.01) \\
    TabM & 0.963 (0.002) & 0.181 (0.005) & 0.226 (0.006) & 0.884 (0.002) & 0.901 (0.007) \\
     + preprocessing & \textbf{0.993 (0.001)} & \textbf{0.186 (0.003)} & 0.578 (0.005) & \textbf{0.887 (0.002)} & \textbf{0.92 (0.008)} \\
    \bottomrule
    \end{tabular}
        \label{tab:preprocessing}
\end{table}

\end{document}